\documentclass[runningheads]{llncs}
\usepackage[T1]{fontenc}
\usepackage{graphicx}
\usepackage{lineno}
\usepackage{hyperref}
\usepackage{paralist}
\usepackage{todonotes}
\usepackage{wrapfig}
\usepackage{subcaption}
\usepackage{booktabs}
\usepackage{array}
\usepackage{caption}
\hypersetup{
    colorlinks=true,
    linkcolor=black,
    citecolor=black,      
    urlcolor=cyan,
    pdfpagemode=FullScreen,
    }
\begin{document}
\title{Real-Time Energy Pricing in New Zealand: \\An Evolving Stream Analysis}
%
%
%
%
%
\author{Yibin Sun\inst{1}\orcidID{0000-0002-8325-1889} \and
Heitor Murilo Gomes\inst{2}\orcidID{0000-0002-5276-637X} \and
Bernhard Pfahringer\inst{1}\orcidID{0000-0002-3732-5787} \and
Albert Bifet\inst{1,3}\orcidID{0000-0002-8339-7773}
}
\authorrunning{Sun et al.}
%
\institute{
University of Waikato, Hamilton, NZ \and
Victoria University of Wellington, Wellington, NZ \and
LTCI, T\'el\'ecom Paris, IP Paris \\
\email{ys388@students.waikato.ac.nz}
}

\maketitle              
\begin{abstract}
This paper introduces a group of novel datasets representing real-time time-series and streaming data of energy prices in New Zealand, sourced from the Electricity Market Information (EMI) website maintained by the New Zealand government. The datasets are intended to address the scarcity of proper datasets for streaming regression learning tasks. We conduct extensive analyses and experiments on these datasets, covering preprocessing techniques, regression tasks, prediction intervals, concept drift detection, and anomaly detection. Our experiments demonstrate the datasets' utility and highlight the challenges and opportunities for future research in energy price forecasting.

\keywords{Data Streams \and Regression \and Prediction Interval \and Anomaly Detection \and Datasets.}
\end{abstract}
\section{Introduction}
Data stream learning has been an active research field over the past two decades. As a promising and powerful tool for big data and AI, streaming algorithms cope with several tasks that are not considered in conventional machine learning.

As outlined in~\cite{ref_moa}, a data stream can arrive at a high velocity with potentially infinite volume. Meanwhile, the distributional properties of streaming data may evolve over time. These issues require streaming algorithms to possess certain characteristics. First, they must digest data points from the data stream efficiently due to the high speed of arrival. Second, they must only read the instances once (or a small number of times) because it is unfeasible to store the large volume of data in memory and iterate as much as desired. Third, they must adapt to Concept Drifts (CD)\cite{ref_CDs}, which is an essential phenomenon in data streams, referring to the property changes over time. Streaming algorithms must be able to detect and adapt to the new concept during the process.

Regression tasks are a crucial subset of machine learning research. However, they are relatively overlooked compared to their classification counterpart, especially in stream learning~\cite{intro_regression}. Only a small number of works have concentrated on streaming regression tasks in recent years~\cite{ref_recentreg1,ref_recentreg2,ref_recentreg3,ref_recentreg4}. The lack of proper datasets is one of the causes.

In this paper, we present a newly introduced real-time streaming dataset, depicting the energy prices in New Zealand, maintained by the \href{https://www.emi.ea.govt.nz}{Electricity Market Information (EMI) website} and provided by the New Zealand government. Extensive analysis and experiments are also conducted on the obtained datasets, affirming their usefulness.

The rest of this paper is organized as follows:
Section~\ref{sec:back} introduces the background information and some related work;
Section~\ref{sec:access} presents the raw data and the recommended preprocessing undertaken for the data;
Section~\ref{sec:experiments} details moderate amount of experiments conducted on the acquired datasets;
Section~\ref{sec:results} exhibits the experimental results; 
and Section~\ref{sec:conclusion} concludes this paper and proposes some possible future work.

\section{Background and Related Work}\label{sec:back}
On 1 November 2022, New Zealand introduced a real-time electricity pricing system, replacing the ex-post final prices system that had been operating since 1 October 1996. This system results in a new price series – Dispatch Energy Prices, which are usually derived from the SPD (Scheduling, Pricing, and Dispatch) model. The prices are summarized and then announced by the pricing manager.

Dispatch Energy Prices are produced throughout trading periods, each of which lasts for 30 minutes. During each trading period, multiple prices are announced and last for one to several minutes. At the end of the trading period, an interim price is calculated as a weighted sum of the produced prices in that period, where the weights are the duration of each price. If no price errors are claimed, the interim price is declared as the final price at 2 PM on the next business day. For more details, please refer to the \href{https://www.emi.ea.govt.nz}{EMI website}.

Due to the vitality of energy and electricity prices to the energy market, research concerning forecasting the energy price has drawn significant attention and produced many solutions. As summarized in \cite{ref_energypricereview1,ref_energypricereview2}, numerous techniques and methodologies have been applied to energy price forecasting tasks over the past decades, many of which are machine learning-based. However, data streams have been absent in this field.

The ML models utilized in this work are introduced as follows.
Sliding Window KNN, Random AMRules~\cite{ref_AMRules}, Adaptive Random Forest Regressor (ARF-Reg)\cite{ref_ARF-Reg}, and Self-Optimising K-Nearest Leaves (SOKNL)\cite{ref_recentreg4}, are advanced algorithms designed for data stream processing and adaptive learning. Sliding Window KNN uses a moving window to handle changing data distributions efficiently. Random AMRules improves rule-based learning by adapting to evolving data patterns. ARF-Reg leverages ensemble learning with random subspace sampling to provide robust regression predictions in dynamic environments. SOKNL integrates KNN and ARF techniques, using leaf centroids for maintaining relevant historical data and adapting to concept drift.
The Adaptive Prediction Interval for Data Stream Regression by Sun~\cite{ref_adapi} dynamically adjusts prediction intervals for accuracy in evolving data streams, using adaptive learning and ensemble techniques to ensure robust predictions.
The Half-Space Trees (HST)\cite{ref_hst} is a method for anomaly detection in data streams, leveraging a forest of trees built through random half-space partitions to identify low-density regions indicative of anomalies. 
\section{Data and Preprocessing}\label{sec:access}
Although EMI provides plenty of different types of data, this work focuses on the Dispatch Energy Prices themselves, leaving other data for future work.
\subsection{Brief Data introduction}\label{sec:dataintro}

The daily dispatch energy prices are available on the website as CSV files.
The files contain the following features:
\begin{itemize}
    \item \texttt{Trading Date}\\the date that the trading takes place in the format of yyyy-mm-dd.
    \item \texttt{Trading Period}\\the trading period indices, usually 1 to 48.
    \item \texttt{Publish Date Time}\\ISO 8601 formatted date and time in NZ of the prices published by the pricing manager.
    \item \texttt{PointOfConnection (PoC)}\\the point of connection on the grid, possesses more than 200 unique values, more details on the \href{https://www.emi.ea.govt.nz/Wholesale/Datasets/MappingsAndGeospatial/NetworkSupplyPointsTable}{Network supply points (NSP) table website}.
    \item \texttt{Island}\\binary values indicating the North Island or South Island of NZ.
    \item \texttt{IsProxyPriceFlag}\\boolean values indicating whether the price is a proxy price, which happens when the PoC is disconnected from the grid due to transmission outage.
    \item \texttt{DollarsPerMegawattHour}\\the published dispatch energy price in dollars per megawatt-hour.
\end{itemize}
All the files are downloadable from the~\href{https://www.emi.ea.govt.nz/Wholesale/Datasets/DispatchAndPricing/DispatchEnergyPrices/}{Dispatch Energy Prices site}.

\subsection{Data Preprocessing}\label{sec:preprocessing}\

\begin{wrapfigure}{R}{.45\textwidth}
    \centering
    \includegraphics[width=.45\textwidth]{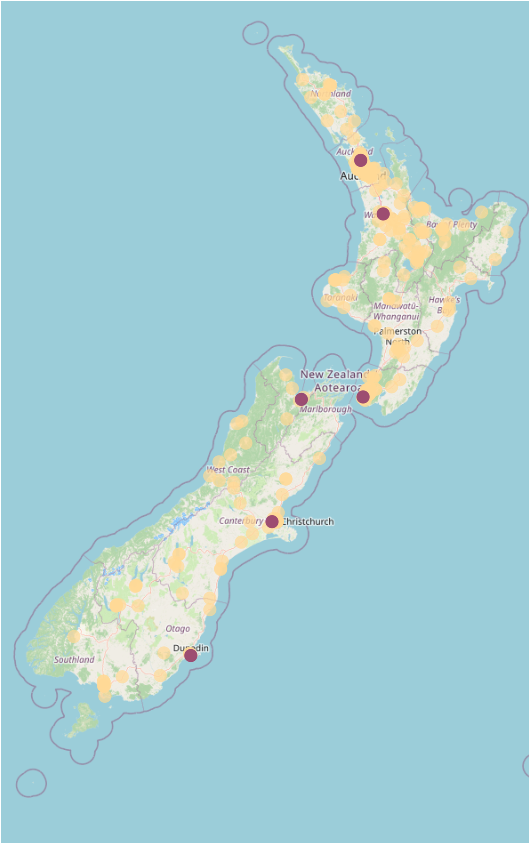}
    \caption{PoCs Overview}
    \label{fig:PoC}
\end{wrapfigure}

As detailed in Section~\ref{sec:dataintro}, most of the attributes included in the data are either time or location related, which are not directly correlated with the energy prices. 
In this work, we propose an example of preprocessing the provided data source for data stream research.
We abstract from 1 Nov. 2022 to 30 Apr. 2024 (18 months) in this work. 

The EMI updates the data files on a daily basis, and it is straightforward to acquire desired information for specified periods utilizing our scripts.
All the associated code is located at our anonymous \href{https://github.com/Anonymous-1-2-3-4/NewZealandEnergyPrices}{Github} repository.

\subsubsection{Regionalization}

Our first focus is the \texttt{PoC} attribute. There are over 200 Point of Connections throughout New Zealand. 
The energy demands and prices are based on the different locations themselves, that says, being close on the geographical level does not necessarily correlate with similar energy prices. 
Therefore, our preprocessing treats each location (PoC) as a separate case.

Figure~\ref{fig:PoC} shows an overview of the PoCs. It is apparent that the PoCs are more dense in the North Island, aligning with the population distribution. 
This work chooses six PoCs, highlighted as purple markers in Figure~\ref{fig:PoC}: Albany, Auckland (\texttt{ALB0331}); Central Hamilton (\texttt{HAM0331}); Wilton, Wellington (\texttt{WIL0331}), Islington, Christchurch (\texttt{ISL0661}), South Dunedin (\texttt{SDN0331}), Stoke, Nelson (\texttt{STK0331}). 
These PoCs are in the main cities in both the North and South Islands of New Zealand, thus representing more of the population. Datasets for other locations are also easily obtainable via the provided code.

\subsubsection{Trading Date and Period}

Trading date and period represent the periodic information in a year and a day, respectively. We convert the date and period index information into numeric values using trigonometric functions.
Specifically, the trading date and period are normalized into [0, $2\pi$] using Equation~\ref{eq:normalization}: 
\begin{equation}
    x_i = 2\pi \cdot \frac{X_i - X_{min}}{X_{max} - X_{min}}
    \label{eq:normalization}
\end{equation}
where $x_i$ is the $i_{th}$ normalized value, $X_i$ is the $i_{th}$ raw value, and $X_{max}$ and $X_{min}$ are the maximum and minimum values in the series, respectively.
Then, the trigonometric functions, Sine and Cosine, are applied to the normalized series. Four new columns are appended to the datasets in this manner: two for the trading period and two for the trading date.

\begin{figure}[h]
    \centering
    \begin{subfigure}[b]{0.48\textwidth}
        \centering
        \includegraphics[width=\textwidth]{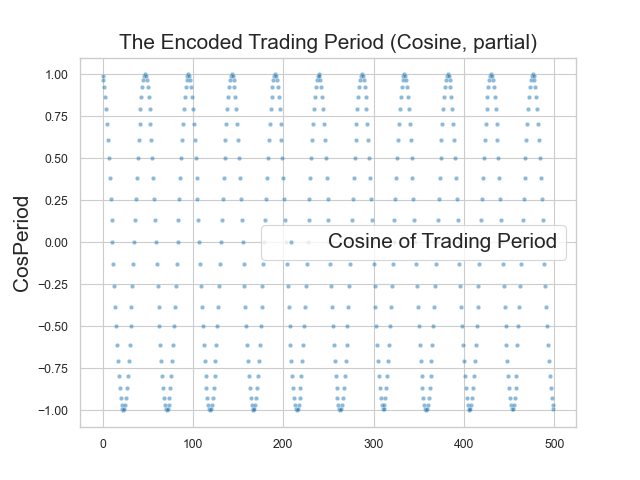}
        \caption{Trading period, showing daily pattern}
        \label{fig:cosperiod}
    \end{subfigure}
    \hfill
    \begin{subfigure}[b]{0.48\textwidth}
        \centering
        \includegraphics[width=\textwidth]{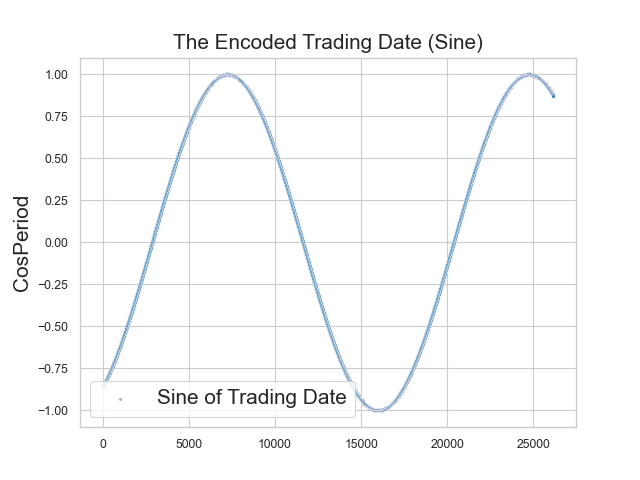}
        \caption{Trading date, showing yearly pattern}
        \label{fig:sindate}
    \end{subfigure}
    \caption{Showcase of the encoded periodical information, PoC: \texttt{HAM0331}.}
    \label{fig:encoded}
\end{figure}

Note that this approach might remove the seasonal information in the original data; therefore, the original information is also retained. Users can select whether they are needed according to the usage context.
Figure~\ref{fig:cosperiod} and~\ref{fig:sindate} illustrate the trading period and date after encoding. 

\subsubsection{Define Targets and Relevant Features}
Under the current NZ energy system, multiple price announcements are made during a single trading period, and the quantity varies for different periods. As a consequence, in our preprocessing phase, we take either the \textit{\textbf{\color{red}mean}} or the \textit{\textbf{\color{red}median}} value to represent a certain trading period as the target.

\begin{figure}[h]
    \centering
    \begin{subfigure}[b]{0.48\textwidth}
        \centering
        \includegraphics[width=\textwidth]{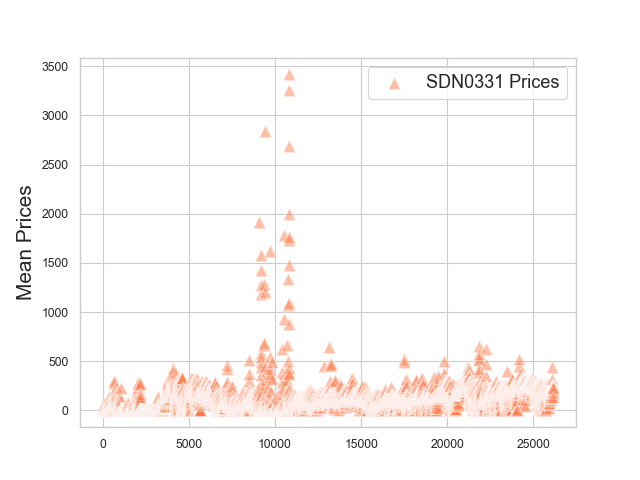}
        \caption{Mean prices from \texttt{SDN0331} PoC.}
        \label{fig:mean}
    \end{subfigure}
    \hfill
    \begin{subfigure}[b]{0.48\textwidth}
        \centering
        \includegraphics[width=\textwidth]{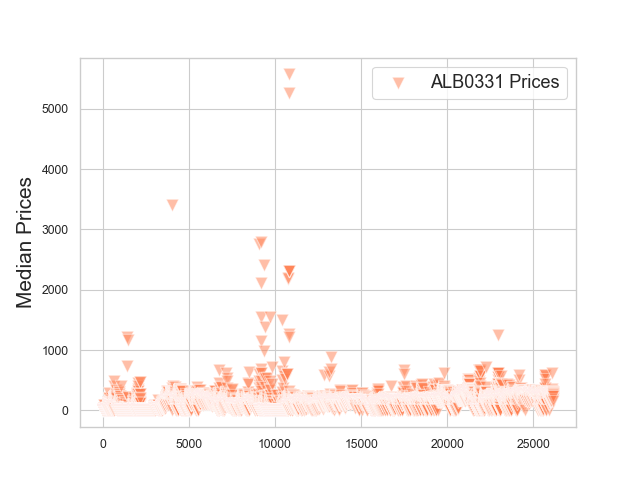}
        \caption{Median prices from \texttt{ALB0331} PoC.}
        \label{fig:median}
    \end{subfigure}
    \caption{Demonstration of target values as mean and median prices.}
    \label{fig:targets}
\end{figure}
Considering the availability of the previously published prices and the high frequency of updates to the database, it is rational and reasonable to include the previous ground truths in the datasets. In our approach, we include both the mean and median values from the past in the feature space. Our scripts allow users to specify how many trading periods need to be predicted in advance, which is defined as \textbf{\textit{\color{red}Delay}} in this article as it is a simulation of the delay of the target values' arrival. We include four groups of datasets, predicting the energy prices 0.5, 4, 6, and 24 hours ahead. By adding these features, we include the time dependence information into the data~\cite{ref_time-dependence}. However, in this article, we focus on data stream mining aspect, leaving other scenarios, such as time-series and auto-regression, for further study.

As observed in Figure~\ref{fig:targets}, the target values, both in mean and median aspects, occasionally reach extremely high levels. These can be defined as outliers or anomalies. Consequently, we also propose datasets for anomaly detection tasks based on the data, where the anomalies are the highest certain percent of values in the datasets. The ratio of the anomalous targets is determined by the users.
\section{Experiments}\label{sec:experiments}
This section introduces experimental settings conducted on the extracted datasets.

\noindent\textbf{Regression.}
As the main purpose and most general use case of this project, regression tasks regarding energy prices are the first assessment in this work. Four state-of-the-art streaming regression algorithms are tested: Sliding Window kNN, Random AMRules (RAMRules)\cite{ref_AMRules}, Adaptive Random Forest for Regression (ARF-Reg)\cite{ref_ARF-Reg}, and Self-Optimising K Nearest Leaves (SOKNL)\cite{ref_recentreg4}. The parameter settings for these algorithms follow the defaults in the Massive Online Analysis (MOA) platform~\cite{ref_moa}, one of the most famous and popular open-source machine learning platforms for data stream learning. All the algorithms used are available in the MOA library. The Coefficient of Determination (R$^2$) is utilized as the evaluation metric in this work for the reason proposed in~\cite{ref_r2better}.

\noindent\textbf{Prediction Interval.}
Prediction Interval (PI)\cite{ref_PIs} is an extension to the regular regression tasks. Instead of a single-valued prediction, PI provides an interval as the prediction, expected to cover a desired ratio of the ground truths.
The prediction interval technique is particularly suitable for energy price forecasting. A predicted range of future prices provides adequate information for decision-making regarding the energy market~\cite{ref_energyPI,ref_energyPI2}.
In this work, we conduct PI tasks with the first proposed, fully incremental, online prediction interval method – Adaptive Prediction Interval (AdaPI)\cite{ref_adapi}. AdaPI adheres to the default settings at~\href{https://capymoa.org}{CapyMOA}, another open-source online learning platform. Only SOKNL is used as the base learner in this work.
Evaluating PI is a challenging task since there are two aspects requiring consideration – the coverage and the interval width. In this work, we exploit \textbf{Coverage} and \textbf{NMPIW} (Normalized Mean Prediction Interval Width) as the evaluation metrics. In the most common cases, Coverage is required to be close to the confidence level while narrowing NMPIW as much as possible~\cite{ref_adapi}.

\noindent\textbf{Drifts.}
Another challenging yet important issue in stream learning is concept drift. In this work, we use Page-Hinckley~\cite{ref_pagehinckley} and Adwin~\cite{ref_adwin} detectors on the Mean Absolute Error (MAE) and Root Mean Squared Error (RMSE) metrics to demonstrate the drifts in the proposed datasets. Both detectors are available on the River online stream learning platform~\cite{ref_river}.

\noindent\textbf{Anomaly Detection.}
As supervised anomaly detection is excessively studied~\cite{ref_surveyAD}, in this paper, we only consider unsupervised streaming anomaly detection tasks. The Half-Space Trees~\cite{ref_hst} algorithm is presented here as a demonstration of the datasets’ usefulness. We use the implementation on the River platform. All experiments are also repeated 10 times with different random seeds.

\section{Results}\label{sec:results}

\subsection{Regression}

Table~\ref{tab:results} illustrates the overall Coefficient of Determination (R$^2$) results from the extensive regression experiments. Ten runs with different random seeds were executed on all datasets and algorithms, and the average R$^2$ values are exhibited in the table. However, all the standard deviation values were smaller than 0.01, so they are omitted in the table.
\begin{table}[!ht]

\setlength{\tabcolsep}{5.2pt} 
    \centering
    \caption{Coefficient of Determination Values by Algorithms, PoCs, and Delays}

    \begin{tabular}{lcccccccc}
        \toprule
       \multicolumn{1}{r}{Alg.}  & \multicolumn{2}{c}{ARF-Reg} & \multicolumn{2}{c}{KNN} & \multicolumn{2}{c}{RAMRules} & \multicolumn{2}{c}{SOKNL} \\
        \cmidrule(r){2-3} \cmidrule(r){4-5} \cmidrule(r){6-7} \cmidrule(r){8-9}
        PoC & Mean & Median & Mean & Median & Mean & Median & Mean & Median \\
        \midrule
        \multicolumn{9}{c}{\texttt{Delay = 30 Minutes}} \\
        \midrule
         \texttt{ALB0331} & 0.56 & 0.54 & 0.6 & 0.57 & 0.2 & 0.22 & 0.61 & 0.58 \\
         \texttt{HAM0331} & 0.57 & 0.54 & 0.6 & 0.58 & 0.25 & 0.31 & 0.61 & 0.59 \\
         \texttt{ISL0661} & 0.63 & 0.62 & 0.65 & 0.64 & 0.59 & 0.59 & 0.67 & 0.65 \\
         \texttt{SDN0331} & 0.64 & 0.64 & 0.67 & 0.66 & 0.6 & 0.61 & 0.69 & 0.67 \\
         \texttt{STK0331} & 0.62 & 0.61 & 0.65 & 0.64 & 0.55 & 0.57 & 0.67 & 0.65 \\
         \texttt{WIL0331} & 0.61 & 0.59 & 0.64 & 0.62 & 0.44 & 0.37 & 0.66 & 0.63 \\
        \midrule
        \multicolumn{9}{c}{\texttt{Delay = 4 Hours}} \\
        \midrule
         \texttt{ALB0331} & 0.5 & 0.48 & 0.5 & 0.47 & 0.4 & 0.38 & 0.57 & 0.54 \\
         \texttt{HAM0331} & 0.51 & 0.49 & 0.51 & 0.47 & 0.42 & 0.41 & 0.58 & 0.55 \\
         \texttt{ISL0661} & 0.56 & 0.55 & 0.57 & 0.54 & 0.49 & 0.48 & 0.63 & 0.62 \\
         \texttt{SDN0331} & 0.58 & 0.57 & 0.58 & 0.56 & 0.52 & 0.5 & 0.65 & 0.63 \\
         \texttt{STK0331} & 0.56 & 0.54 & 0.56 & 0.54 & 0.48 & 0.46 & 0.63 & 0.61 \\
         \texttt{WIL0331} & 0.55 & 0.53 & 0.55 & 0.51 & 0.43 & 0.43 & 0.62 & 0.59 \\
        \midrule
        \multicolumn{9}{c}{\texttt{Delay = 6 Hours}} \\
        \midrule
         \texttt{ALB0331} & 0.51 & 0.48 & 0.49 & 0.45 & 0.34 & 0.24 & 0.57 & 0.54 \\
         \texttt{HAM0331} & 0.51 & 0.48 & 0.5 & 0.46 & 0.39 & 0.25 & 0.58 & 0.55 \\
         \texttt{ISL0661} & 0.56 & 0.55 & 0.56 & 0.54 & 0.5 & 0.49 & 0.63 & 0.61 \\
         \texttt{SDN0331} & 0.58 & 0.57 & 0.58 & 0.56 & 0.52 & 0.51 & 0.65 & 0.63 \\
         \texttt{STK0331} & 0.56 & 0.54 & 0.55 & 0.53 & 0.49 & 0.48 & 0.63 & 0.6 \\
         \texttt{WIL0331} & 0.55 & 0.53 & 0.54 & 0.5 & 0.38 & 0.32 & 0.62 & 0.59 \\
        \midrule
        \multicolumn{9}{c}{\texttt{Delay = 24 Hours}} \\
        \midrule
         \texttt{ALB0331} & 0.53 & 0.49 & 0.46 & 0.43 & 0.43 & 0.4 & 0.58 & 0.56 \\
         \texttt{HAM0331} & 0.53 & 0.5 & 0.47 & 0.44 & 0.47 & 0.41 & 0.59 & 0.57 \\
         \texttt{ISL0661} & 0.57 & 0.56 & 0.53 & 0.51 & 0.57 & 0.55 & 0.63 & 0.63 \\
         \texttt{SDN0331} & 0.59 & 0.58 & 0.55 & 0.52 & 0.55 & 0.56 & 0.65 & 0.65 \\
         \texttt{STK0331} & 0.57 & 0.56 & 0.53 & 0.5 & 0.57 & 0.56 & 0.63 & 0.62 \\
         \texttt{WIL0331} & 0.57 & 0.54 & 0.51 & 0.48 & 0.51 & 0.5 & 0.63 & 0.61 \\
\bottomrule
\end{tabular}
\label{tab:results}
\end{table}

Key takeaways of Table~\ref{tab:results}:
\begin{inparaenum}[(a)]
    \item \texttt{ISL0661}, \texttt{SDN0331}, \texttt{STK0331}, and \texttt{WIL0331} demonstrate relatively stable R$^2$ values across different algorithms, while \texttt{ALB0331} and \texttt{HAM0331} show more fluctuation.
    \item SOKNL consistently produces better R$^2$ results for all PoCs. RAMRules performs less competitively in most cases. 
    \item As expected, the delay length has a negative impact on the performance of all algorithms. Whereas, the performance drops are at an acceptable level.
\end{inparaenum}

\begin{figure}
    \centering
    \includegraphics[width=\textwidth]{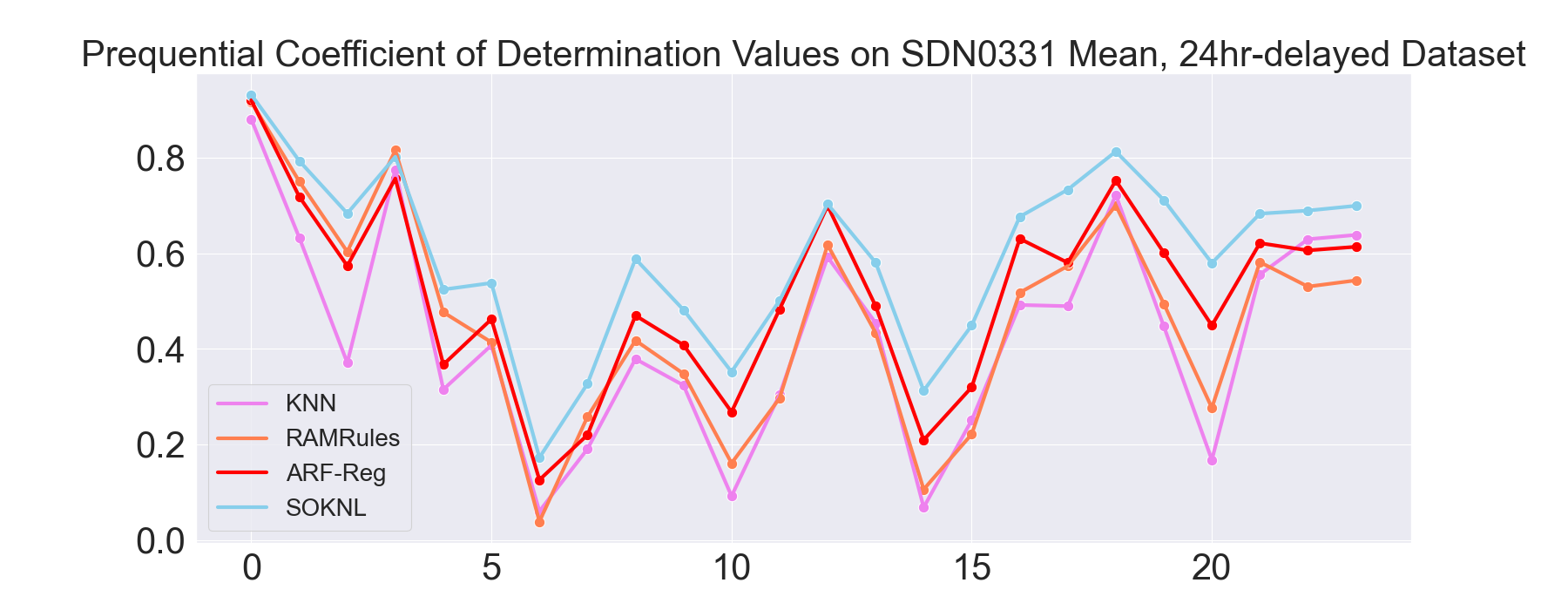}
    \caption{\texttt{SDN0331} PoC: Prequential R$^2$ Score with 1000 Window Size}
    \label{fig:prequential}
\end{figure}

Figure~\ref{fig:prequential} illustrates an example exhibition of prequential R$^2$ results. Due to the space limit, only \texttt{SDN0331} PoC with mean values as the targets is displayed here. The performance for four algorithms fluctuates with a similar tendency, seemingly affirming several concept drifts in the data. 

\subsection{Prediction Interval}
\begin{table}[!]
\centering
\caption{Coverage and NMPIW values for different PoCs and Delays}
\begin{tabular}{lcccccccc}
\toprule
\multicolumn{1}{r}{Delay} &        \multicolumn{2}{c}{30 Minutes}    &           \multicolumn{2}{c}{4 Hours} &   \multicolumn{2}{c}{6 Hours} &          \multicolumn{2}{c}{24 Hours} \\
PoC     &    Coverage & NMPIW           &       Coverage & NMPIW         &         Coverage & NMPIW       &        Coverage & NMPIW        \\
\midrule
\multicolumn{9}{c}{\texttt{Confidence Level = 95\%}}\\
\midrule
\texttt{ALB0331} &  98.03 & 4.91 &   97.90 & 5.18 &   97.90 & 5.18 &   97.87 & 5.10 \\
\texttt{HAM0331} &  98.02 & 4.88 &  97.86 & 5.16 &   97.82 & 5.20 &  97.82 & 5.12 \\
\texttt{ISL0661} &  97.26 & 4.97 &  97.12 & 5.28 &  97.14 & 5.26 &  97.18 & 5.21 \\
\texttt{SDN0331} &  97.20 & 5.04 &  97.03 & 5.36 &  96.99 & 5.34 &  97.12 & 5.30 \\
\texttt{STK0331} &  97.33 & 4.91 &  97.16 & 5.22 &  97.14 & 5.19 &  97.24 & 5.18 \\
\texttt{WIL0331} &  97.32 & 4.88 &  97.12 & 5.21 &  97.18 & 5.22 &  97.14 & 5.12 \\

\midrule
\multicolumn{9}{c}{\texttt{Confidence Level = 80\%}}\\
\midrule
\texttt{ALB0331} &   93.60 & 2.98 &  92.68 & 3.18 &  92.58 & 3.18 &  92.56 & 3.12 \\
\texttt{HAM0331} &  93.48 & 2.97 &  92.52 & 3.17 &   92.36 & 3.20 &  92.44 & 3.14 \\
\texttt{ISL0661} &  90.66 & 3.01 &  89.72 & 3.21 &   89.70 & 3.22 &   89.80 & 3.19 \\
\texttt{SDN0331} &  90.52 & 3.04 &  89.57 & 3.26 &  89.36 & 3.25 &  89.53 & 3.24 \\
\texttt{STK0331} &  90.91 & 2.98 &  89.81 & 3.16 &  89.78 & 3.18 &  89.98 & 3.16 \\
\texttt{WIL0331} &  91.65 & 2.96 &  90.34 & 3.18 &  90.46 & 3.19 &  90.44 & 3.13 \\

\bottomrule
\end{tabular}
\label{table:coverage_nmpiw}
\end{table}

Table~\ref{table:coverage_nmpiw} presents the PI results with 80\% and 95\% confidence level. 

Overall, the coverage values are all higher than the desired confidence level, and the NMPIW values are at a low level. This means that for these datasets, AdaPI can generate relatively compact intervals that cover more-than-required ground truths. This can greatly benefit decision-making in the energy market.

For each PoC, both Coverage and NMPIW values remain relatively stable across different delays, indicating that the model’s prediction accuracy and interval width do not significantly fluctuate with the delay.
There is also a clear trade-off between coverage and NMPIW, i.e., higher coverage usually requires wider NMPIW. 
Among the PoCs, ALB0331 and HAM0331 generally show higher coverage and relatively lower NMPIW compared to other PoCs.

\begin{figure}[!ht]
    \centering
    \includegraphics[width=\textwidth]{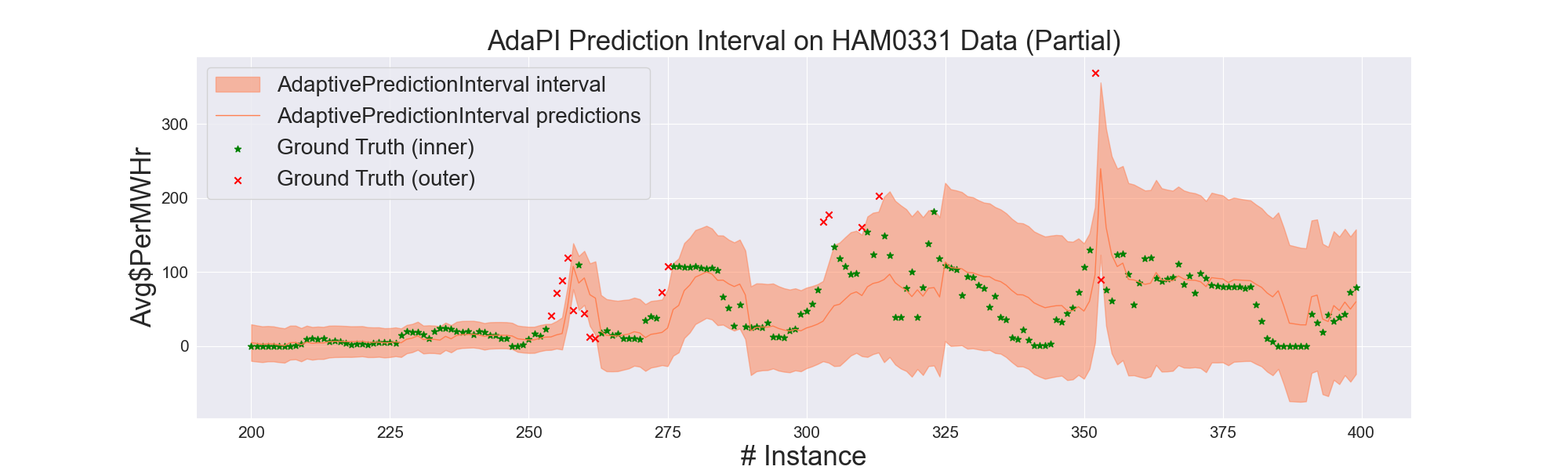}
    \caption{Visualized Example of Prediction Intervals Over Time}
    \label{fig:pi_overtime}
\end{figure}
Figure~\ref{fig:pi_overtime} is a visualization of the real-time generated prediction intervals. Only a small portion of the whole dataset is presented here. It can be seen that the data starts fluctuating around 250 instances, and AdaPI adapts to it by enlarging the PI area to protect the coverage from decreasing dramatically. 

\subsection{Drifts}
\begin{figure}[!ht]
    \centering
    \begin{minipage}{\textwidth}
        \subfloat[RMSE from KNN Regression by Page-Hinckley with 30 Minutes Delay]{
            \includegraphics[width=.95\textwidth]{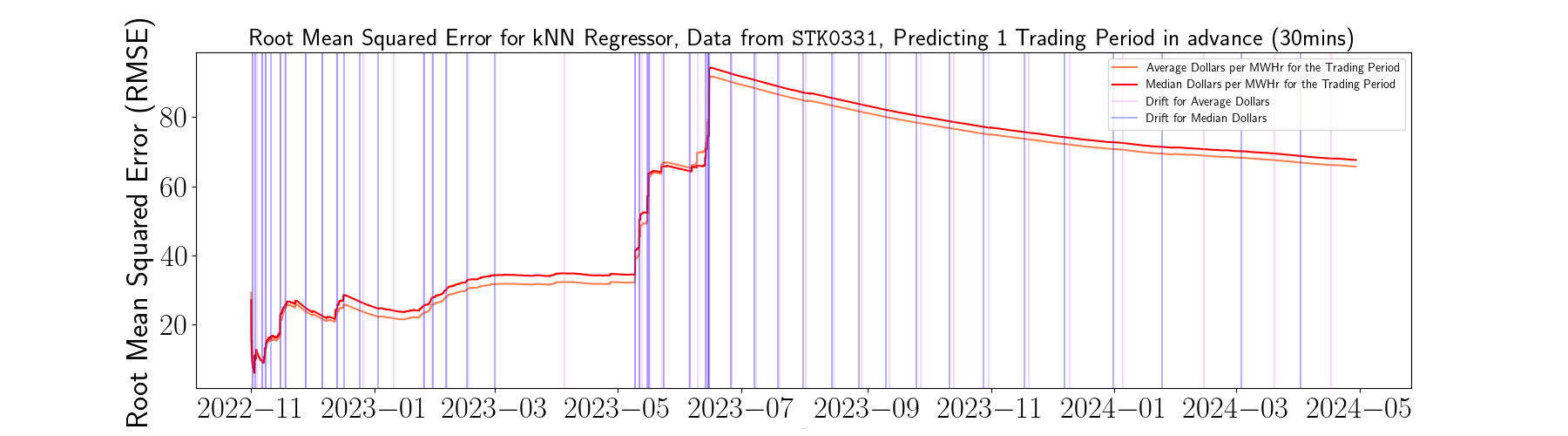}
            \label{fig:knn_drift}
        }
    \end{minipage}
    \vfill
    \begin{minipage}{\textwidth}
        \subfloat[MAE from SOKNL by ADWIN with 24 Hours Delay]{
            \includegraphics[width=.95\textwidth]{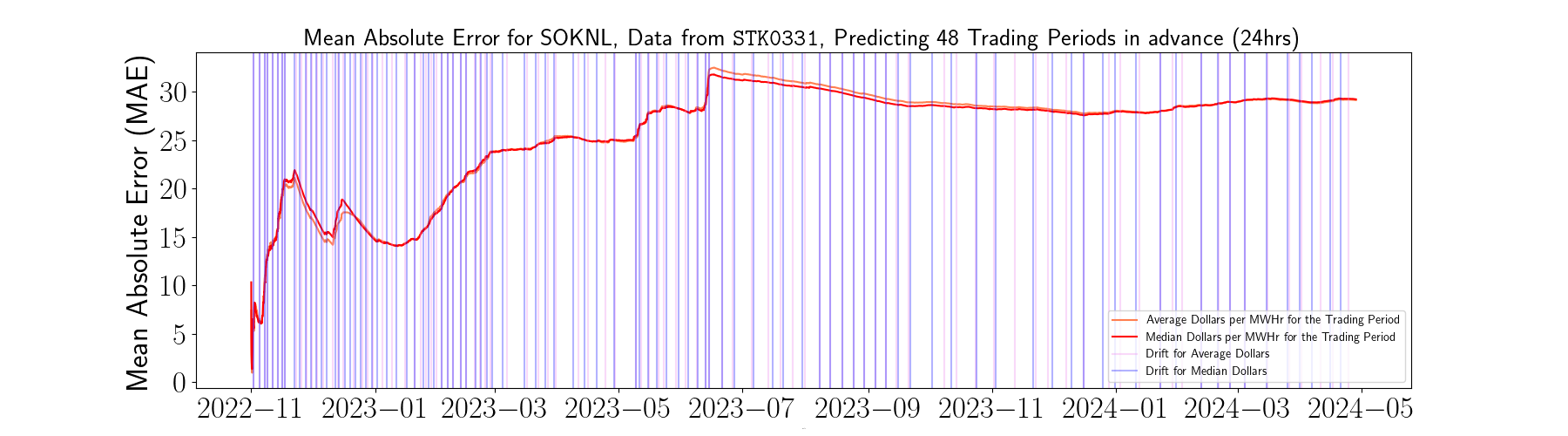}
            \label{fig:soknl_drift}
        }
    \end{minipage}
    \caption{Drift Analysis Showcases on Data from \texttt{STK0331}}
    \label{fig:drifts}
\end{figure}

Figure~\ref{fig:drifts} illustrates examples of drift analysis on data from the \texttt{STK0331} PoC. Among them, Figure~\ref{fig:knn_drift} includes the drifts detected by the Page-Hinckley test based on the RMSE values from the KNN regressor. Figure~\ref{fig:soknl_drift}, on the other hand, highlights the drifts detected by an ADWIN detector using MAE from the SOKNL algorithm. Both median and average targets are involved in the figures. The orange line and violet vertical lines represent the performance and the detected drifts on the average dataset, respectively. Similarly, the red line and blue vertical lines represent the errors and drifts on the median dataset.

One can see that drifts happen quite frequent in the energy prices. This justifies the reason to study and analyse these datasets using data stream techniques instead of traditional batch learning ones. 

There are more drift reported in Figure~\ref{fig:soknl_drift} than in Figure~\ref{fig:knn_drift} because ADWIN is more sensitive than the Page-Hinckley test. There are two apparent error increases between 2023-05 and 2023-07, which align with the winter season in New Zealand. Intuitively, they are related; however, more investigation is needed for confirmation.

\subsection{Anomaly Detection}
\begin{table}[!ht]

\setlength{\tabcolsep}{5.2pt} 
    \centering
    \caption{Half-Space Trees AUC Scores by Anomaly Ratios, PoCs, and Delays}

    \begin{tabular}{lcccccccc}
        \toprule
        \multicolumn{1}{r}{Ratio$_A$} & \multicolumn{2}{c}{0.1\%} & \multicolumn{2}{c}{0.5\%} & \multicolumn{2}{c}{1\%} & \multicolumn{2}{c}{2\%} \\
        \cmidrule(r){2-3} \cmidrule(r){4-5} \cmidrule(r){6-7} \cmidrule(r){8-9}
        PoC & Mean & Median & Mean & Median & Mean & Median & Mean & Median \\
        \midrule
        \multicolumn{9}{c}{\texttt{Delay = 30 Minutes}} \\
        \midrule
         \texttt{ALB0331} & 0.597 & 0.668& 0.472 & 0.446 & 0.429 & 0.421 & 0.454 & 0.452 \\
         \texttt{HAM0331} & 0.598 & 0.669 & 0.47 & 0.45 & 0.425 & 0.419 & 0.442 & 0.445 \\
         \texttt{ISL0661} & 0.573 & 0.629 & 0.468 & 0.486 & 0.504 & 0.519 & 0.5 & 0.503 \\
         \texttt{SDN0331} & 0.541 & 0.627 & 0.484 & 0.497 & 0.501 & 0.507 & 0.518 & 0.525 \\
         \texttt{STK0331} & 0.572 & 0.628 & 0.471 & 0.461 & 0.527 & 0.541 & 0.496 & 0.5 \\
         \texttt{WIL0331} & 0.611 & 0.65 & 0.469 & 0.462 & 0.433 & 0.451 & 0.456 & 0.454 \\
        \midrule
        \multicolumn{9}{c}{\texttt{Delay = 4 Hours}} \\
        \midrule
         \texttt{ALB0331} & 0.388 & 0.446 & 0.382 &0.375 & 0.38 & 0.381 & 0.428 & 0.435 \\
         \texttt{HAM0331} & 0.388 & 0.446 & 0.383 & 0.378 & 0.378 & 0.379 & 0.418 & 0.428 \\
         \texttt{ISL0661} & 0.342 & 0.386 & 0.378 & 0.423 & 0.461 & 0.486 & 0.474 & 0.484 \\
         \texttt{SDN0331} & 0.313 & 0.386 & 0.404 & 0.436 & 0.461 & 0.474 & 0.492 & 0.506 \\
         \texttt{STK0331} & 0.343 & 0.386 & 0.386 & 0.402 & 0.485 & 0.505 & 0.471 & 0.482 \\
         \texttt{WIL0331} & 0.389 & 0.428 & 0.377 & 0.391 & 0.386 & 0.413 & 0.429 & 0.437 \\
        \midrule
        \multicolumn{9}{c}{\texttt{Delay = 6 Hours}} \\
        \midrule
         \texttt{ALB0331} & 0.387 & 0.444 & 0.384 & 0.377 & 0.377 & 0.383 & 0.424 & 0.433 \\
         \texttt{HAM0331} & 0.388 & 0.444 & 0.384 & 0.379 & 0.375 & 0.38 & 0.413 & 0.426 \\
         \texttt{ISL0661} & 0.340 & 0.380 & 0.378 & 0.428 & 0.465 & 0.491 & 0.472 & 0.483 \\
         \texttt{SDN0331} & 0.312 & 0.379 & 0.402 & 0.438 & 0.461 & 0.475 & 0.49 & 0.504 \\
         \texttt{STK0331} & 0.34 & 0.38 & 0.383 & 0.406 & 0.482 & 0.504 & 0.468 & 0.48 \\
         \texttt{WIL0331} & 0.39 & 0.426 & 0.378 & 0.393 & 0.383 & 0.413 & 0.427 & 0.435 \\
        \midrule
        \multicolumn{9}{c}{\texttt{Delay = 24 Hours}} \\
        \midrule
         \texttt{ALB0331} & 0.565 & 0.571 & 0.416 & 0.405 & 0.414 & 0.414 & 0.439 & 0.443 \\
         \texttt{HAM0331} & 0.565 & 0.571 & 0.416 & 0.403 & 0.414 & 0.414 & 0.433 & 0.44 \\
         \texttt{ISL0661} & 0.524 & 0.513 & 0.463 & 0.491 & 0.526 & 0.543 & 0.498 & 0.511 \\
         \texttt{SDN0331} & 0.494 & 0.512 & 0.488 & 0.503 & 0.542 & 0.554 & 0.524 & 0.54 \\
         \texttt{STK0331} & 0.523 & 0.513 & 0.468 & 0.47 & 0.522 & 0.534 & 0.496 & 0.501 \\
         \texttt{WIL0331} & 0.562 & 0.546 & 0.404 & 0.404 & 0.426 & 0.461 & 0.458 & 0.464 \\
\bottomrule
\end{tabular}
\label{tab:ADresults}
\end{table}

Table~\ref{tab:ADresults} exhibits the Receiver Operating Characteristic - Area Under the Curve (ROC AUC) scores from Half-Space Trees algorithm executed on the processed datasets with different anomaly ratios.

Overall, these datasets appear quite challenging for Half-Space Trees as most of the AUC scores are under 0.5. Due to the temporal characteristics of the datasets, we did not apply a shuffle procedure to the data. Additionally, in most of the datasets, the anomalies (high values of price) are not distributed in a consistent pattern. As illustrated in Figure~\ref{fig:targets}, the outliers are concentrated in a specific time period (possibly May -- July 2023, taking into account Figure~\ref{fig:drifts}). This may cause difficulty for the streaming anomaly detection algorithms.

More specifically, the performance of Half-Space Trees is influenced by both the delay and the anomaly ratio. It performs best with shorter delays and lower anomaly ratios. However, with longer delays, it adapts and performs better with higher anomaly ratios. The regression tasks can benefit from the anomaly detection to determine if an “abnormal” target value should be treated as an anomaly or the start of a drift. To align with the data stream gist, we did not consider offline anomaly detection. Further research is required in this respect.

\section{Conclusion}\label{sec:conclusion}
In this study, we presented a comprehensive analysis of a newly introduced real-time streaming dataset of energy prices in New Zealand. Our experiments with various state-of-the-art streaming regression algorithms and prediction interval techniques demonstrated the dataset’s potential for accurate and reliable energy price forecasting. The dataset also proved valuable for detecting concept drifts and anomalies, providing a robust foundation for further research in these areas. Despite the promising results, our findings also highlight several challenges, such as the need for better handling of extreme outliers and the impact of seasonal variations on prediction accuracy. Future work will focus on improving preprocessing techniques, exploring additional machine learning models, and enhancing the dataset’s applicability to other regions and energy markets.
Future work will try to introduce meteorological information in the datasets.

\bibliographystyle{abbrv}
\bibliography{references}
\end{document}